\documentclass[letterpaper]{article} 
\usepackage{aaai2026}  
\usepackage{times}  
\usepackage{helvet}  
\usepackage{courier}  
\usepackage[hyphens]{url}  
\usepackage{graphicx} 
\usepackage{natbib}  
\usepackage{caption} 
\usepackage{algorithm}
\usepackage{algorithmic}

\usepackage{newfloat}
\usepackage{listings}
\DeclareCaptionStyle{ruled}{labelfont=normalfont,labelsep=colon,strut=off} 
\floatstyle{ruled}
\newfloat{listing}{tb}{lst}{}
\floatname{listing}{Listing}
\usepackage{multirow}
\usepackage{makecell}
\usepackage{xcolor}
\usepackage{bm}
\usepackage{diagbox}
\usepackage{amssymb}
\usepackage{enumitem}
\usepackage{amsmath}
\usepackage{booktabs}

\newcommand{\f}[1]{\textbf{#1}}
\newcommand{\s}[1]{#1}

\title{Composition-Incremental Learning for Compositional Generalization}
\author{
    Zhen Li\textsuperscript{\rm 1,2},
    Yuwei Wu\textsuperscript{\rm 1,2},
    Chenchen Jing\textsuperscript{\rm 3},
    Che Sun\textsuperscript{\rm 2}\thanks{Corresponding author: Che Sun and Chuanhao Li},
    Chuanhao Li\textsuperscript{\rm 4,1,2}\footnotemark[1],
    Yunde Jia\textsuperscript{\rm 2,1}
}
\affiliations{
    \textsuperscript{\rm 1}Beijing Key Laboratory of Intelligent Information Technology,\\School of Computer Science \& Technology, Beijing Institute of Technology\\
    \textsuperscript{\rm 2}Guangdong Laboratory of Machine Perception and Intelligent Computing, Shenzhen MSU-BIT University\\
    \textsuperscript{\rm 3}Zhejiang University of Technology\\
    \textsuperscript{\rm 4}Shanghai AI Laboratory\\
    \{li.zhen, wuyuwei\}@bit.edu.cn, jingchenchen@zju.edu.cn, \{sunche, jiayunde\}@smbu.edu.cn, lichuanhao@pjlab.org.cn
}

\begin{document}

\maketitle

\begin{abstract}
Compositional generalization has achieved substantial progress in computer vision on pre-collected training data.
Nonetheless, real-world data continually emerges, with possible compositions being nearly infinite, long-tailed, and not entirely visible.
Thus, an ideal model is supposed to gradually improve the capability of compositional generalization in an incremental manner.
In this paper, we explore \textbf{Comp}osition-\textbf{I}ncremental \textbf{L}earning for Compositional Generalization \mbox{(CompIL)} in the context of the compositional zero-shot learning (CZSL) task, where models need to continually learn new compositions, intending to improve their compositional generalization capability progressively.
To quantitatively evaluate CompIL, we develop a benchmark construction pipeline leveraging existing datasets, yielding MIT-States-CompIL and C-GQA-CompIL.
Furthermore, we propose a pseudo-replay framework utilizing a visual synthesizer to synthesize visual representations of learned compositions and a linguistic primitive distillation mechanism to maintain aligned primitive representations across the learning process.
Extensive experiments demonstrate the effectiveness of the proposed framework.
\end{abstract}

\begin{links}\textbf{- This is a copy of the copyrighted version at AAAI.}\end{links}

\section{Introduction}

\begin{figure}[t]
    \centering
    \includegraphics[width=1.0\linewidth]{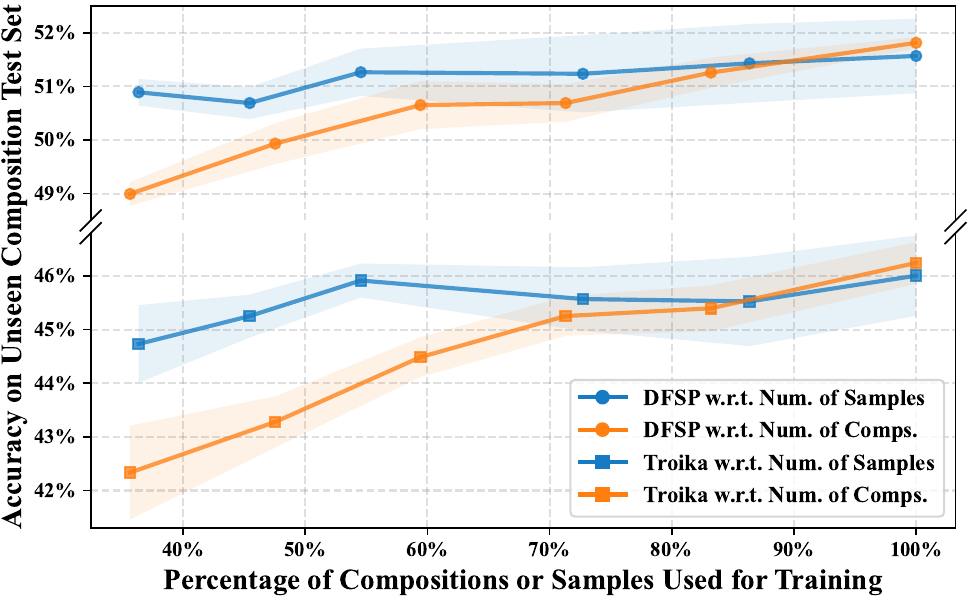}
    \caption{Accuracy of CZSL models on the unseen composition test set trained with data containing varying number of compositions or samples.
    Increasing training compositions boosts compositional generalization significantly more than increasing training samples.}
    \label{fig:motivation}
\end{figure}

Recently, compositional generalization has garnered much attention, with substantial progress in improving models' compositional generalization capability on fixed, pre-collected data~\cite{huang2024troika,huang2024towards,li2023exploring}.
Given the ever-emerging nature of real-world data, \emph{e.g.}, the recurrence of previously observed compositions and the appearance of new, unseen ones, it is essential to understand how to enrich the training data to further boost the compositional generalization capability of models.
To this end, we conduct a preliminary investigation into the impact of training data on this capability in the compositional zero-shot learning (CZSL) task, which aims to recognize unseen compositions of attributes and objects (known as primitives) by leveraging knowledge from observed compositions.
Specifically, we conduct comparative experiments by varying the number of samples in the training data while keeping the number of compositions fixed, or vice versa.
As illustrated in Figure~\ref{fig:motivation}, the steeper slope of the orange line indicates that the number of compositions has a significantly impact on models' compositional generalization capability, while increasing sample size with a fixed number of compositions offers minimal benefit.
More details can be found in the \textbf{supplementary material}.
These findings suggest that we can improve the compositional generalization capability of models by increasing the diversity of training compositions from a data-driven perspective.

Nevertheless, obtaining sufficiently varied compositions is both costly and time-consuming~\cite{saini2024beyond,xu2024mac}, and the expense of training models from scratch becomes prohibitive as the dataset size scales. This raises an important question: \textbf{\textit{Can models continually learn from an increasing number of compositions to improve their compositional generalization capability progressively?}}
To answer this question, we propose a new setting: Composition-Incremental Learning for Compositional Generalization (CompIL), where models are required to learn sequentially on a series of tasks containing disjoint compositions.
Specifically, each task contains a set of samples sharing the same primitive set, whereas the compositions of different tasks vary significantly in semantics.
These semantics gaps simulate the staged data collection process in the real world, where data distribution typically follows a long-tailed pattern and continuously evolves~\cite{gama2014survey, yao2022wild,li2024searchlvlms}.
The difficulty of our setting stems from the following two challenges:
(1) \textbf{Composition knowledge forgetting.}
Forgetting~\cite{li2017learning} remains a fundamental challenge in continual learning and is even more pronounced in our setting.
The vast number of compositions, coupled with the relatively small number of samples per composition, exacerbates the risk of forgetting.
(2) \textbf{Primitive representation drift.}
Learning semantically aligned primitive representations has been proven to enhance the compositional generalization capability of models~\cite{li2023exploring}.
However, the semantic differences between tasks foster models focusing on task-specific representations, which may not be applicable across tasks.
For example, in one task, ``ancient'' emphasizes age, as in ``ancient castle'', while in another, it emphasizes obsolescence, as in ``ancient computer''.

We explore CompIL in the context of CZSL, and develop an efficient benchmark construction pipeline along with a comprehensive 
evaluation protocol.
By formalizing the constraints in benchmark construction as an integer optimization problem, our pipeline constructs CompIL benchmarks from existing datasets via step-by-step optimization.
Moreover, we introduce a hierarchical clustering strategy to enhance inter-task semantic diversity, enabling the benchmark to better align with the dynamic real world.
In practice, we construct two new benchmarks MIT-States-CompIL and C-GQA-CompIL. We evaluate various existing continual learning methods and find they struggle on CompIL, often underperforming even a zero-shot baseline.

We present a pseudo-replay framework for CompIL by synthesizing pseudo-samples of past compositions and training them jointly with current task data.
Recognizing that the compositions is infinite and difficult to disentangle in visual representations~\cite{lu2023decomposed}, while primitives are naturally separable in language (\emph{e.g.}, attribute and object words), we design a visual synthesizer based on the language encoder of a pretrained vision-language model.
Leveraging the vision-and-language alignment of the pretrained model, the synthesizer learns to take attribute and object words as input and synthesize corresponding visual composition representations, as pseudo-samples.
Additionally, we introduce a linguistic primitive distillation mechanism. 
It constrains the model to maintain consistent predictions for past compositions while learning new ones, effectively mitigating primitive representation drift.
Experimental results show our framework consistently improves the compositional generalization capability of models throughout the learning process.

To summarize, our contributions are as follows: 
\begin{itemize}[leftmargin=2.0em]
    \item We present a practical and challenging setting termed CompIL, where models continually learn new compositions to improve their compositional generalization capability progressively. 
    \item We develop an efficient pipeline for constructing CompIL benchmarks for quantitative evaluation and construct two new benchmarks in the context of CZSL.
    \item We propose a pseudo-replay framework for CompIL by synthesizing visual representations of learned compositions and maintaining aligned primitive representations throughout learning. 
\end{itemize}

\section{Related Work}

\begin{figure*}[t]
    \centering
    \includegraphics[width=1\linewidth]{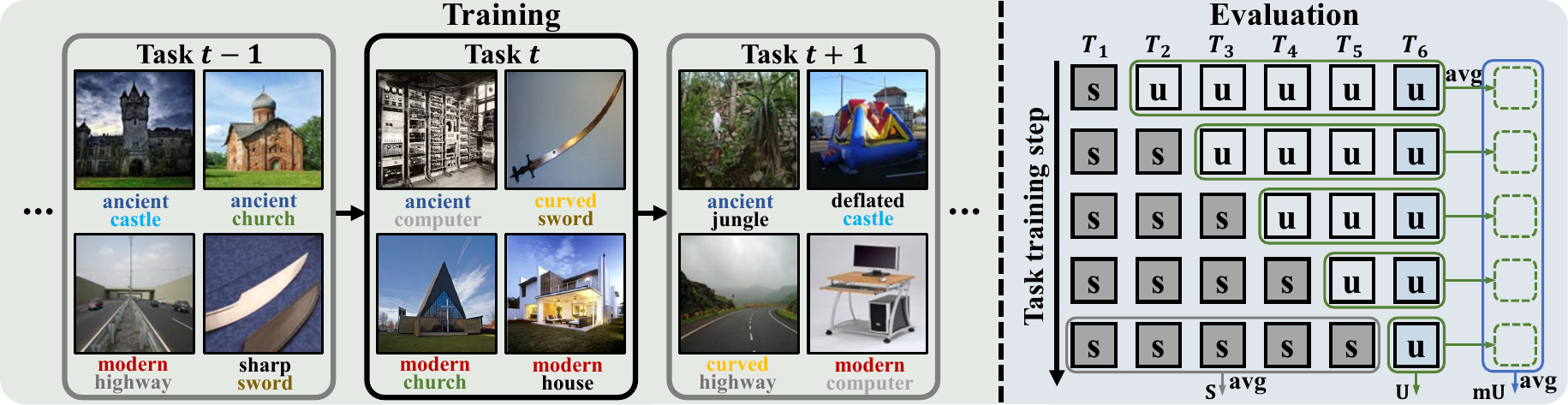}
    \caption{
    Illustration of our CompIL setting, taking the proposed MIT-States-CompIL benchmark as an example.
    The left side illustrates training samples from the tasks, where samples from different tasks vary significantly in semantics.
    For instance, the attribute ``ancient'' describes buildings in task $t-1$, outdated technology in task $t$, and natural landscapes in task $t+1$; the object ``castle'' varies likewise.
    The right side depicts the evaluation process, with rows representing training steps and each column indicating performance on the corresponding task.
    }
    \label{fig:benchmark}
\end{figure*}

\subsection{Compositional Generalization}

Numerous benchmarks~\cite{ma2023crepe,li2024compositional,ray2024cola} have been proposed to evaluate compositional generalization capability, and various sophisticated model architectures and training strategies~\cite{huang2024towards,li2024context} have been proposed to boost this capacity.
A key area of compositional generalization research is compositional zero-shot learning.
Benefiting from the capability of pre-trained vision-language models, \emph{e.g.}, CLIP~\cite{radford2021learning}, diverse cross-model mechanisms have been proposed to enhance compositional generalization capability.
For example,  replacing attributes and object labels with trainable prompts~\cite{nayak2023learning}, employing cross-modal fusion to enhance feature integration~\cite{lu2023decomposed}, and using multi-branch models to better align vision-language representations~\cite{huang2024troika}.
These works focus on improving the compositional generalization capability on pre-collected and fixed data. Differently, our work aims to improve this capability progressively using a growing data stream with various compositions to cope with the ever-changing world.

A few works have explored extending the boundaries of compositional generalization with increasing data.
VisCOLL~\cite{jin-etal-2020-visually} investigated the incremental acquisition of compositional phrases from streaming visual data and evaluated the compositional generalization capability after the learning process.
\citet{liao2024does} focused on the multi-object compositions and proposed a compositional few-shot testing protocol for evaluating compositional generalization in continual learning.
CCZSL~\cite{ijcai2024p191} required models to continually learn compositions that include unseen primitives to expand the learned primitive set over time.
CompILer~\cite{zhangnot} also introduced a composition-incremental learning task, which separately identifies attributes and objects, aiming to mitigate forgetting of each.
In contrast to the above, we explore learning continually from an increasing number of compositions within a fixed primitive set, aiming to improve models' compositional generalization capability progressively.

\subsection{Continual Learning}

Continual learning is to train a single model that can incrementally update its knowledge with a continuous stream of tasks without catastrophic forgetting of previously learned tasks.
Existing methods alleviated catastrophic forgetting via regularization~\cite{kirkpatrick2017overcoming,dhar2019learning}, expanding models for each task~\cite{li2019learn,hu2023dense}, or storing samples of previous tasks~\cite{chaudhry2019continual,buzzega2020dark,li2024towards}. 

Recently, several works have shown interest in continual learning with CLIP.
\citet{thengane2022clip} showed that CLIP achieves state-of-the-art performance via a zero-shot paradigm in continual learning settings. 
AttriCLIP~\cite{wang2023attriclip} leveraged a trainable attribute word bank to encode image attributes as textual prompts, enabling efficient continual learning while mitigating catastrophic forgetting.
CGIL~\cite{frascaroli2024clip} trained a dedicated Variational Autoencoder~\cite{kingma2013auto} for each class to generate synthetic visual features that are then used for the continual adaptation of CLIP models.
Although these methods have demonstrated impressive results in mitigating catastrophic forgetting or preventing zero-shot capability degradation, their use to enhance compositional generalization capability remains under-explored.
By contrast, we propose a pseudo-replay framework based on visual composition synthesis to enhance compositional generalization capability while mitigating forgetting.

\section{Composition-Incremental Learning}

\begin{figure*}[t]
    \centering
    \includegraphics[width=1.0\linewidth]{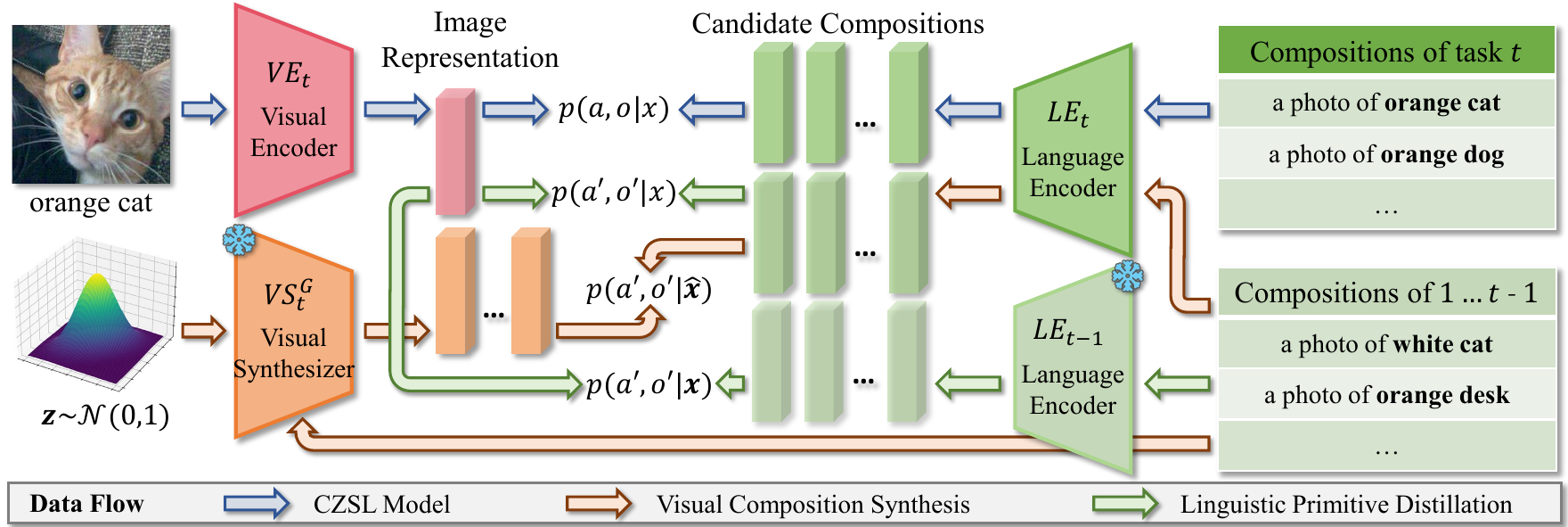}
    \caption{
    Overview of the pseudo-replay framework in the context of compositional zero-shot learning.
    The CZSL model contains a visual encoder $\mathit{VE}_t$ and a language encoder $\mathit{LE}_t$. For simplicity, we omit possible connections between two components.
    }
    \label{fig:framework}
\end{figure*}

\subsection{Formulation}

We take CZSL as a representative example to illustrate the formulation of CompIL.
Given an attribute set $\mathcal{A} = \{a_1, a_2, \dots, a_{|\mathcal{A}|}\}$ and an object set $\mathcal{O} = \{o_1, o_2, \dots, o_{|\mathcal{O}|}\}$ as the primitive concepts, the compositional label space $\mathcal{C} = \mathcal{A} \times \mathcal{O}$ is defined as their Cartesian product.
CZSL divides $\mathcal{C}$ into 2 disjoint subsets, \emph{i.e.}, $\{\mathcal{C}_1, \mathcal{C}_2\}$, aiming at learning a model from $\mathcal{C}_1$ to recognize images from novel composition set $\mathcal{C}_2$.
In composition-incremental learning, the composition set $\mathcal{C}$ is further divided into $T+ 1$ disjoint subsets, \emph{i.e.}, $\{\mathcal{C}_1, \mathcal{C}_2, \dots, \mathcal{C}_T, \mathcal{C}_{T + 1}\}$, where $\mathcal{C}_{i} \cap \mathcal{C}_{j} = \emptyset$ for any $i \ne j$ and $\bigcup_{t=1}^{T+1} \mathcal{C}_t \subseteq \mathcal{C}$.
Each subset $\mathcal{C}_t$ (except for $\mathcal{C}_{T+1}$) with corresponding images $\mathcal{X}_t$ forms a task of CompIL, denoted as $\mathcal{T}_t = \{(x_i, c_i) | x \in \mathcal{X}_t, c \in \mathcal{C}_t\}$, resulting in a total of $T$ tasks for the continual learning process.

The model is trained sequentially across these tasks.
When learning the $t$-th task, the training images only contain compositions from $\mathcal{C}_t$, while the evaluation is performed on both seen composition set $\mathcal{C}_t^s$ and unseen composition set $\mathcal{C}_t^u$ following the standard compositional zero-shot learning, where $\mathcal{C}_t^s = \bigcup_{i=1}^{t} \mathcal{C}_i$ and $\mathcal{C}_t^u = \bigcup_{i=t+1}^{T+1} \mathcal{C}_i$, respectively. 
Note that $\mathcal{C}_{T+1}$ is always included in the unseen set $\mathcal{C}_t^u$ for any task $t$, providing a static unseen composition set for consistent and quantitative evaluation of the model's compositional generalization capability during the learning process. 

\subsection{Evaluation Metric}

The evaluation encompasses two aspects: the model’s average performance throughout the continual learning process and its final performance after the process, as illustrated in the right part of Figure~\ref{fig:benchmark}.
Specifically, after training on task $t$, we follow the well-established CZSL evaluation protocol by \citet{purushwalkam2019task} to evaluate the model on the seen composition set $\mathcal{C}_t^s$ and the unseen composition set $\mathcal{C}_t^u$, including the best seen accuracy $\textit{S}_t$, the best unseen accuracy $\textit{U}_t$, and the area under the curve $\textit{AUC}_t$ for unseen versus seen accuracy.
We report the average best unseen accuracy $\textit{mU}=\frac{1}{T} \sum_{i=1}^{T} \textit{U}_i$ and the average area under the curve $\textit{mAUC}=\frac{1}{T}{\sum_{i=1}^{T} \textit{AUC}_i}$, which measures the model’s capability when continually learning new tasks.
The best seen accuracy $\textit{S}_t$ can be further divided by task as $\textit{S}_{t,i}$ 
to quantify the performance degradation in the past tasks, which is defined as $\textit{fS} = {\frac{1}{T} \sum_{i=1}^{T} (\textit{S}_{i, i} - \textit{S}_{T, i}})$.
Additionally, we report the final $\textit{U}_T$, $\textit{S}_T$ and $\textit{AUC}_T$, denoted as $\textit{U}$, $\textit{S}$ and $\textit{AUC}$, to reflect the model’s performance after the learning process.

\subsection{Benchmark Construction}

To quantitatively assess the performance of models in CompIL, we propose a pipeline that constructs CompIL benchmarks leveraging existing datasets.
Besides ensuring the formulation, the pipeline also simulates the staged data collection process in the real world, where data distribution typically follows a long-tailed pattern and evolves over time.
Specifically, semantically similar compositions tend to be densely observed within a period, leading to semantic differences between tasks.

The pipeline employs a hierarchical clustering strategy to take the aforementioned considerations into account.
Given an existing CZSL dataset containing a set of compositions and corresponding images, we first cluster the compositions into $M$ semantically similar mini-groups $\{\mathcal{G}_1, \mathcal{G}_2, \dots, \mathcal{G}_M\}$, using K-Means algorithm~\cite{lloyd1982least}.
Each mini-group consists of compositions with the same attribute and semantically similar objects.
The semantic similarity is quantified using Lin similarity~\cite{lin1998information} calculated on WordNet~\cite{miller1995wordnet}.
Next, we assign each mini-group to one of the $T$ tasks.
Considering the definition of compositional generalization, which refers to unseen compositions of seen primitives, we obtain a shared primitive set across different tasks by maximizing the objective function
\begin{equation}
    \sum_{t=1}^{T} 
    \left[ 
        \mathrm{N_a} \left( \bigcup_{i=1}^{M} \mathcal{G}_i \mid \mathcal{G}_i \in \mathcal{T}_t \right) + 
        \mathrm{N_o} \left( \bigcup_{i=1}^{M} \mathcal{G}_i \mid \mathcal{G}_i \in \mathcal{T}_t \right)
    \right],
\end{equation}
where $\mathrm{N_a}(\cdot)$ and $\mathrm{N_o}(\cdot)$ are functions that calculate the number of attribute and object types given the composition set, respectively. 
The objective function can be transformed into an integer optimization problem, and we use Gurobi Optimizer~\cite{gurobi} to find an approximate solution in a finite number of optimization steps.
Finally, the unseen composition test set of the CZSL dataset is designated as task $T+1$.
We combine it with the above $T$ tasks to form a CompIL benchmark.

We use the pipeline construct MIT-States-CompIL and C-GQA-CompIL benchmarks containing $T=5$ tasks based on widely used CZSL datasets MIT-States~\cite{isola2015discovering} and C-GQA~\cite{mancini2022learning}.
Taking the MIT-States-CompIL benchmark as an example, Figure~\ref{fig:benchmark} illustrates our CompIL setting.
Additional statistics can be found in \textbf{supplementary material}.

\section{Pseudo-Replay Framework}

The overview of the proposed framework in the context of compositional zero-shot learning is shown in Figure \ref{fig:framework}.
Concretely, for a CZSL model containing a visual encoder $\mathit{VE}_t$ and a language encoder $\mathit{LE}_t$, the framework integrates a visual synthesizer $\mathit{VS}$ that synthesizes visual representations of past tasks.
The synthesized representations are combined with current task samples to train the CZSL model jointly.
Additionally, the language encoder of the CZSL model, finalized on the last task, is preserved and utilized to perform distillation with the current language encoder.

\subsection{Preliminary}

We first outline the pipeline of recent mainstream CZSL methods leveraging the pretrained vision-language model, \emph{i.e.}, CLIP~\cite{radford2021learning}.
When training on task $t$, given an input image $x$ and a candidate composition set $\mathcal{C}_t$, these methods take a visual encoder $\mathit{VE}_t$ and a language encoder $\mathit{LE}_t$ to obtain the image representation $\bm{x} = \mathit{VE}_t(x)$ and candidate compositions representations $\{\bm{c} = \mathit{LE}_t(a, o)| (a, o) \in \mathcal{C}_t\}$, respectively.
Both $\mathit{VE}_t$ and $\mathit{LE}_t$ are based on CLIP~\cite{radford2021learning}, and candidate composition representations are generated using prompt templates like ``a photo of \{attribute\} \{object\}''.
Then, the recognition probability of the input image is calculated as
\begin{equation}
p(a,o|x) = \frac{\exp(\cos(\mathit{VE}_t(x), \mathit{LE}_t(a, o)) / \tau)}{\sum_{i=1}^{|\mathcal{C}_t|}\exp(\cos(\mathit{VE}_t(x), \mathit{LE}_t(a_i, o_i)) / \tau)}, 
\end{equation}
where $\tau$ denotes the temperature, $\cos(\cdot, \cdot)$ is the cosine similarity function, and $(a, o)$ is the target composition.
On this basis, diverse mechanisms have been proposed, such as learnable prompts~\cite{nayak2023learning}, cross-modal interaction modules~\cite{huang2024troika}, and retrieval augmentation modules~\cite{jing2024retrieval}, to further enhance the compositional generalization capability of the model.

\subsection{Visual Composition Synthesis}

\begin{figure}[t]
    \centering
    \includegraphics[width=1.0\linewidth]{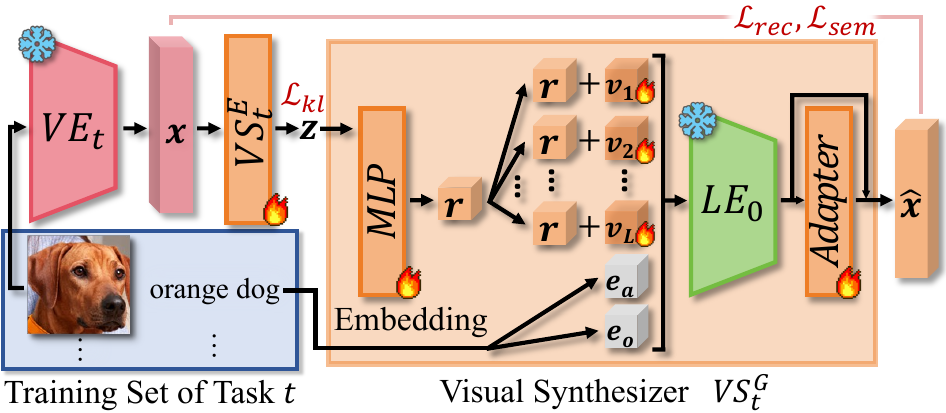}
    \caption{The architecture and training of the visual synthesizer. The visual synthesizer comprises an encoder $\mathit{VS}^E_t$ and a generator $\mathit{VS}^G_t$. It is trained to synthesize visual representations conditioned on the given composition name.}
    \label{fig:framework-vs}
\end{figure}

We propose a visual synthesizer that learns to synthesize visual representations of past tasks.
The visual synthesizer employs the Variational Autoencoder~\cite{kingma2013auto} architecture, comprising an encoder $\mathit{VS}^E_t$ and a generator $\mathit{VS}^G_t$, as illustrated in Figure~\ref{fig:framework-vs}.
The encoder $\mathit{VS}^E_t$, implemented as a simple fully-connected network, encodes the image representation $\bm{x}$ into a latent code $\bm{z}$. The generator $\mathit{VS}^G_t$ synthesizes image representations using the latent code $\bm{z}$ and corresponding attribute and object name $(a, o)$.

Following \citet{wang2023improving}, we adapt the language encoder $\mathit{LE}_0$ (\emph{i.e.}, the language encoder of pretrained CLIP) for the generator $\mathit{VS}_G$, aiming to enhance the learning efficiency and the quality of the synthesizer by leveraging the aligned vision and language representations learned by the pretrained CLIP.
Thus, given the latent code $\bm{z}$ and the composition $(a, o)$, instead of synthesizing the image representation directly, the generator learns to synthesize instance-specific prompts
\begin{equation}
\bm{p}(\bm{z}, a, o) = [\bm{v}_1 + \bm{r}, \bm{v}_2 + \bm{r}, ..., \bm{v}_L + \bm{r}, \bm{e}_a, \bm{e}_o], 
\end{equation}
where $\{\bm{v}_1, \bm{v}_2, ..., \bm{v}_L\}$ are learnable prompts of length $L$, $\bm{e}_a$ and $\bm{e}_o$ are token embedding of the corresponding attribute and object $(a, o)$, $\bm{r}$ is the local bias obtained from the latent code $\bm{z}$ through a fully-connected network.
Then, the prompts are fed into the language encoder $\mathit{LE}_0$ to obtain the synthesized image representation $\hat{\bm{x}}$.
Additionally, a lightweight adapter~\cite{gao2024clip} is introduced to further bridge the modality gap.
Thus, given an image representation $\bm{x}$, the synthesis process is described as
\begin{equation}
\bm{z} = \mathit{VS}^E_t(\bm{x}), \hat{\bm{x}} = \mathit{VS}^G_t(\bm{z}, a, o) = \mathit{LE}_0(\bm{p}).
\end{equation}
The optimization of the visual synthesizer is achieved via a standard evidence-lower bound
\begin{equation}
\mathcal{L}_{rec} = || \bm{x} - \hat{\bm{x}}||_2, \mathcal{L}_{kl} = \text{KL}(\bm{z}, \mathcal{N}(0, 1)),
\end{equation}
where $\text{KL}$ is the Kullback-Leibler divergence.

To ensure semantic consistency, we minimize the difference between the primitive semantic distributions of the synthesized and original visual representations.
These distributions are computed via similarity to candidate primitives using simple prompt templates (\emph{e.g.}, “an object looks {attribute}”). The attribute semantic distribution is denoted as
\begin{equation}
p(a|\bm{x}) = \frac{\exp(\cos(\bm{x}, \mathit{LE}_0(a)) / \tau)}{\sum_{i=1}^{|\mathcal{A}|}\exp(\cos(\bm{x}, \mathit{LE}_0(a_i)) / \tau)},
\end{equation}
and the object semantic distribution $p(o|\bm{x})$ is conducted similarly. Thus, the semantic loss is calculated as
\begin{equation}
\mathcal{L}_{sem} = \text{KL}(p(a|\hat{\bm{x}}), p(a|\bm{x})) + \text{KL}(p(o|\hat{\bm{x}}), p(o|\bm{x})).
\end{equation}
overall optimization objective of the visual synthesizer is
\begin{equation}
\mathcal{L}_{VS} = \mathcal{L}_{rec} + \mathcal{L}_{kl} + \alpha \cdot\mathcal{L}_{sem},
\end{equation}
where $\alpha$ is the hyper-parameter that balances the objective of element-wise reconstruction and the semantic consistency.

\begin{table*}[t]
    \small
    \centering
    \begingroup
    \setlength{\tabcolsep}{3.5pt}
    \renewcommand{\arraystretch}{0.80}
    \begin{tabular}{l|cccccc|cccccc}
    \toprule[1.2pt]
    \multirow{2.5}{*}{\diagbox[width=0.27\textwidth]{Method}{Baseline}} & \multicolumn{6}{c|}{CSP~\cite{nayak2023learning} (CLIP ViT-L/14)} & \multicolumn{6}{c}{Troika~\cite{huang2024troika} (CLIP ViT-B/16)} \\
    \cmidrule(lr){2-7}\cmidrule(lr){8-13}
    & $\textit{U}$ & $\textit{S}$ & $\textit{AUC}$ & $\textit{mU}$ & $\textit{fS}(\downarrow)$ & $\textit{mAUC}$ & $\textit{U}$ & $\textit{S}$ & $\textit{AUC}$ & $\textit{mU}$ & $\textit{fS}(\downarrow)$ & $\textit{mAUC}$ \\
    \midrule[0.5pt]
    Zero-Shot                           & 46.10 & 30.63 & 11.15 & -     & -     & -     & 41.02 & 28.15 & 8.95 & -    & -     & -     \\
    Joint                               & 49.61 & 46.51 & 19.24 & -     & -     & -     & 47.47 & 44.12 & 17.35 & -    & -     & -     \\
    \midrule[0.5pt]
    Vanilla                             & 39.78 & 30.55 & 9.76 & 41.41 & 15.22 & 13.59 & 36.84 & 26.98 & 7.54 & 38.40 & 26.69 & 10.68 \\
    SI~\cite{zenke2017continual}        & 46.88 & 37.40 & 14.33 & 45.96 & 8.30 & 16.45 & 36.69 & 27.52 & 7.54 & 38.42 & 26.50 & 10.70 \\
    EWC~\cite{chaudhry2019continual}    & \s{48.35} & \s{39.62} & \s{15.70} & \s{46.81} & 7.21 & \s{17.30} & \s{41.39} & 29.71 & 9.41 & \s{40.70} & 21.62 & 11.80 \\
    \midrule[0.5pt]
    A-GEM~\cite{chaudhry2018efficient}  & 45.71 & 36.22 & 13.46 & 45.71 & 10.17 & 16.32 & 39.54 & 28.87 & 8.67 & 39.71 & 24.31 & 11.34 \\
    DER++~\cite{buzzega2020dark}        & 45.73 & 37.31 & 13.92 & 43.28 & \s{6.19} & 14.72 & 39.39 & \s{30.50} & \s{9.20} & 38.74 & \f{8.46} & 10.59 \\
    \midrule[0.5pt]
    L2P~\cite{wang2022learning}         & 38.92 & 30.50 & 9.41 & 40.39 & 15.25 & 13.17 & 37.42 & 27.65 & 7.95 & 37.97 & 23.72 & 10.58 \\
    AttriCLIP~\cite{wang2023attriclip}  & 39.12 & 30.67 & 9.65 & 42.15 & 15.69 & 14.17 & 37.27 & 26.93 & 7.64 & 37.36 & 27.69 & 10.05 \\
    \midrule[0.5pt]
    \textbf{Ours}                       & \f{49.11} & \f{41.43} & \f{16.77} & \f{47.23} & \f{3.28} & \f{17.99} & \f{42.60} & \f{32.40} & \f{10.69} & \f{41.59} & \s{10.72} & \f{12.26} \\
    \bottomrule[1.2pt]
    \end{tabular}
    \endgroup
    \caption{Comparison with state-of-the-art continual learning methods on CZSL models on the MIT-States-CompIL benchmark.}
    \label{tab:mit-states}
\end{table*}

\subsection{Linguistic Primitive Distillation}

Learning semantically aligned primitives improves compositional generalization~\cite{li2023exploring}.
In order to encourage the model to learn primitive semantic representations applicable to all previously seen compositions, rather than overfitting to the compositions of the current task, the framework incorporates a distillation-based mechanism.
Specifically, after completing the training on task $t - 1$, the language encoder $\mathit{LE}_{t-1}$ is duplicated and frozen.
During the training of task $t$, for a given image representation $\bm{x}$ (whether it is derived from an image of the current task encoded by the vision encoder $\mathit{VE}_t$ or synthesized by the vision synthesizer $\mathit{VS}$), the predicted logits $\bm{l}_t$ using the current $\mathit{LE}_t$ for all compositions from past tasks are computed as
\begin{equation}
\bm{l}_t = \{\cos(\bm{x}, \mathit{LE}_t(a_i, o_i)) / \tau \ | \ (a_i, o_i) \in \bigcup\nolimits_{j=1}^{t} \mathcal{C}_j \}.
\end{equation}
Similarly, the logits $\bm{l}_{t-1}$ can be computed using the duplicated language encoder $\mathit{LE}_{t-1}$.
The distillation loss
\begin{equation}
\mathcal{L}_{kd} = || \bm{l}_t - \bm{l}_{t-1}||_2
\end{equation}
encourages consistent predictions for past compositions, ensuring that updates for new tasks retain previously learned aligned primitive representations.

Notably, benefiting from our visual synthesizer, the visual representations can correspond to any composition from task $0$ to task $t$.
This enables meaningful linguistic primitive distillation across all past tasks, highlighting the difference from vanilla knowledge distillation.
Such distillation facilitates the learning of unified primitive semantic representations that generalize across tasks.
Furthermore, these unified linguistic primitives ensure the synthesized visual features remain semantically aligned throughout different training stages.
In other words, the visual composition synthesis and the linguistic primitive distillation can promote each other.

\subsection{Optimization}

To incorporate a CZSL model into the proposed framework, we begin by duplicating the language encoder of the model and utilizing it as part of the visual synthesizer.
Subsequently, the model is initially trained on task $0$ using the method-specific loss $\mathcal{L}_\mathit{ms}$
, which depends on the selected CZSL model.
After completing training on task $t-1$, we duplicate and freeze the language encoder $\mathit{LE}_{t-1}$ of the CZSL model. 
Then, the visual synthesizer is optimized on the training set of task $t-1$ with the objective $\mathcal{L}_\mathit{VS}$.
The encoder of the synthesizer $\mathit{VS}^E_t$ is randomly initialized per task, while $\mathit{VS}^G_t$ is retained and updated across tasks.
When training begins on task $t$, the synthesizer is employed to synthesize visual representations of past compositions, given the composition name $(a', o')$ and the noise $\bm{z}$ sampled from the prior distribution $\mathcal{N}(0, 1)$.
These synthesized representations $\hat{\bm{x}}$ are combined with the current task's training samples to train the CZSL model using the method-specific loss $\mathcal{L}_\mathit{ms}$.
The overall optimization objective of the model is
\begin{equation}
\mathcal{L} = \mathcal{L}_\mathit{ms} + \beta \cdot \mathcal{L}_{kd},
\end{equation}
where the hyper-parameter $\beta$ balances the stability-plasticity trade-off in primitive representation learning.

\section{Experiments}

\subsection{Experiment Setting}

\noindent
\textbf{Baseline Models.}  We apply the proposed framework to two CZSL models, CSP~\cite{nayak2023learning} and Troika~\cite{huang2024troika}.
CSP adopts the CLIP model with learnable prompts in the language encoder, similar to the common paradigms in class-incremental learning, where the visual encoder is frozen, and only the classification head is trained.
In contrast, Troika employs a more complex network architecture with additional trainable parameters, \emph{e.g.}, the cross-modal interaction modules, reflecting the cutting-edge advancements in CZSL.
We implement the two models with pre-trained CLIP ViT-L/14 and ViT-B/16 to validate the proposed framework across different model scales. More results are in the \textbf{supplementary material}.
Notably, for models that require patch features during training, \emph{e.g.}, Troika, we repeat the synthesized representations to match the size of the patch features and use them for joint training.

\noindent
\textbf{Implementation Details.}
The learning rate is set to 1e-4 for all experiments.
We halve the training epochs in the original paper to prevent overfitting on individual tasks: CSP is trained for 10 epochs per task on MIT-States-CompIL and C-GQA-CompIL, while Troika is trained for 5 and 7 epochs, respectively.
The prompt length $L$ is set to 3.
The hyper-parameter $\alpha$ and $\beta$ are set to 0.1 and 0.3.
All experiments are run three times under different random seeds, and the average results are reported.

\noindent
\textbf{Comparison Methods.} Since CompIL is a newly proposed setting, there does not exist any prior works that can be used for comparison directly.
Therefore, we reimplement and adapt three types of continual learning methods to integrate them with CZSL models for fair comparison.
These include regularization-based approaches SI~\cite{zenke2017continual} and EWC~\cite{chaudhry2019continual}, rehearsal-based methods A-GEM~\cite{chaudhry2018efficient}, and DER++~\cite{buzzega2020dark}, and the recent prompt-based approaches L2P~\cite{wang2022learning} and AttriCLIP~\cite{wang2023attriclip}.
The memory size of rehearsal-based methods is set to 5\% of the total training samples.
We conduct hyperparameter search for these methods to ensure fair comparison, and provide additional results (\emph{e.g.}, buffer sizes) in the \textbf{supplementary material}.

\begin{table}[t]
    \setlength{\aboverulesep}{1.6pt}
    \setlength{\belowrulesep}{1.6pt}
    \small
    \centering
    \begingroup
    \setlength{\tabcolsep}{5pt}
    \renewcommand{\arraystretch}{1.05}
    \begin{tabular}{l|ccccccc}
        \toprule[1.2pt]
        \multirow{2.5}{*}{Method} & \multicolumn{7}{c}{Task Number} \\
        \cmidrule(lr){2-8}
        & 0 & 1 & 2 & 3 & 4 & 5 & Avg \\
        \midrule[0.5pt]
        CSP                         & 10.2 & \f{6.2} & 8.3 & 1.9 & 4.2 & 3.2 & 5.6 \\
        \textbf{CSP + Ours}         & \f{10.2} & 6.0 & \f{8.4} & \f{2.4} & \f{6.3} & \f{3.8} & \f{6.2}\\
        \midrule[0.5pt]
        Troika                      & 16.4 & 12.1 & \f{23.4} & 10.9 & 18.9 & 10.7 & 15.4 \\
        \textbf{Troika + Ours}      & \f{17.2} & \f{13.3} & 22.5 & \f{14.0} & \f{22.2} & \f{15.6} & \f{17.5} \\
        \bottomrule[1.2pt]
    \end{tabular}
    \endgroup
    \caption{Results of the proposed framework on the split of the C-GQA dataset introduced in CCZSL. We reported the AUC in each task and the average AUC across all tasks.}
    \label{tab:cczsl-c-gqa}
\end{table}

\begin{table*}[t]
    \small
    \centering
    \begingroup
    \setlength{\tabcolsep}{4.75pt}
    \renewcommand{\arraystretch}{0.80}
    \begin{tabular}{l|cccccc|cccccc}
        \toprule[1.2pt]
        \multirow{2.5}{*}{\diagbox[width=0.27\textwidth]{Method}{Baseline}} & \multicolumn{6}{c|}{CSP (CLIP ViT-L/14)} & \multicolumn{6}{c}{Troika (CLIP ViT-B/16)} \\
        \cmidrule(lr){2-7}\cmidrule(lr){8-13}
        & $\textit{U}$ & $\textit{S}$ & $\textit{AUC}$ & $\textit{mU}$ & $\textit{fS}(\downarrow)$ & $\textit{mAUC}$ & $\textit{U}$ & $\textit{S}$ & $\textit{AUC}$ & $\textit{mU}$ & $\textit{fS}(\downarrow)$ & $\textit{mAUC}$ \\
        \midrule[0.5pt]
        Zero-Shot                           & 25.18 & 7.39 & 1.41 & -     & -     & -     & 23.95 & 6.75 & 1.16 & -    & -     & -     \\
        Joint                               & 27.80 & 28.88 & 6.35 & -     & -     & -     & 34.18 & 42.87 & 12.63 & -    & -     & -     \\
        \midrule[0.5pt]
        Vanilla                             & 15.39 & 16.62 & 1.89 & 19.28 & 8.86 & 3.86 & 22.73 & 27.31 & 4.95 & 24.97 & 22.54 & 9.70 \\
        SI~\cite{zenke2017continual}        & 18.27 & 18.61 & 2.53 & 20.12 & 6.32 & 4.64 & 22.38 & 28.23 & 5.00 & 24.39 & 21.46 & 9.56 \\
        EWC~\cite{chaudhry2019continual}    & \s{24.48} & \f{21.42} & \s{3.91} & \s{24.25} & \f{3.69} & \s{5.10} & \s{26.05} & 29.26 & \s{6.33} & \s{27.17} & 21.50 & \s{11.17} \\
        \midrule[0.5pt]
        A-GEM~\cite{chaudhry2018efficient}  & 17.66 & 18.69 & 2.53 & 20.45 & 7.06 & 4.36 & 23.51 & 26.83 & 5.01 & 23.90 & 23.52 & 9.11 \\
        DER++~\cite{buzzega2020dark}        & 22.64 & \s{21.24} & 3.59 & 22.40 & \s{4.25} & 4.64 & 21.33 & \s{32.85} & 5.84 & 20.00 & \f{8.32} & 8.45 \\
        \midrule[0.5pt]
        L2P~\cite{wang2022learning}         & 18.27 & 19.02 & 2.62 & 17.22 & 9.29 & 3.78& 21.24 & 28.93 & 4.72 & 23.01 & 20.77 & 9.65 \\
        AttriCLIP~\cite{wang2023attriclip}  & 19.76 & 19.90 & 2.99 & 21.80 & 12.52 & 5.05 & 23.08 & 26.45 & 4.90 & 23.62 & 24.26 & 10.00 \\
        \midrule[0.5pt]
        \textbf{Ours}                        & \f{27.54} & 20.81 & \f{4.60} & \f{26.71} & 5.67 & \f{5.42} & \f{29.55} & \f{34.57} & \f{8.40} & \f{27.88} & \s{14.61} & \f{11.98} \\
        \bottomrule[1.2pt]
    \end{tabular}
    \endgroup
    \caption{Comparison with state-of-the-art continual learning methods on CZSL models on the C-GQA-CompIL benchmark.}
    \label{tab:c-gqa}
\end{table*}

\subsection{Results on Composition-Incremental Setting}

The experimental results on  MIT-States-CompIL and C-GQA-CompIL are listed in Table \ref{tab:mit-states} and \ref{tab:c-gqa}, where ``Zero-Shot'' refers to predictions from the pretrained CLIP model, ``Joint'' (upper bound) represents training all tasks jointly, and ``Vanilla'' (lower bound) represents simply performing gradient update task by task.
We observe that:
(1) Our framework consistently enhances two CZSL models on both seen and unseen compositions, achieving state-of-the-art overall performance as measured by $\textit{AUC}$ and $\textit{mAUC}$.
(2) Our framework significantly improves compositional generalization ($\textit{U}$ and $\textit{mU}$), surpassing the second-best method by an average of 2\%, while effectively mitigating forgetting ($\textit{S}$ and $\textit{fS}$).
(3) Existing methods struggle to continually improve compositional generalization, mostly performing worse than or similar to Zero-Shot after the learning process (see $\textit{U}$ metrics).  
Besides, prompt-based methods (L2P and AttriCLIP) fail in CompIL. We speculate this arises from conflicts between prompt learning and the updates of other learnable parameters in the baseline models.

\subsection{Results on Primitive-Incremental Setting}

Unlike our CompIL setting that focuses on composition-incremental learning within a fixed primitive set, CCZSL~\cite{ijcai2024p191} requires models to continuously learn from compositions that include unseen primitives, thereby expanding the size of the learned primitive set over time.
We conduct experiments on the split of the C-GQA~\cite{mancini2022learning} dataset introduced in CCZSL, and the experimental results are shown in Table \ref{tab:cczsl-c-gqa}.
We observe that our framework improves CSP and Troika across different sessions on the CCZSL split of the C-GQA dataset, with 0.6\% and 2.1\% absolute gains in the average AUC.
Such observations suggest that our framework effectively improves CZSL models on primitive-incremental settings, though it is not explicitly designed for that.

\begin{table}[t]
    \small
    \centering
    \begingroup
    \setlength{\tabcolsep}{5pt}
    \renewcommand{\arraystretch}{0.8}
    \begin{tabular}{lccc|>{\centering\arraybackslash}p{35pt}>{\centering\arraybackslash}p{35pt}>{\centering\arraybackslash}p{35pt}}
        \toprule[1.2pt]
        \ & $\mathit{VS}$ & $\mathcal{L}_{sem}$ & $\mathcal{L}_{kd}$ & $\textit{mU}$ & $\textit{fS}(\downarrow)$ & $\textit{mAUC}$ \\
        \midrule[0.5pt]
        1 & \           & \             & \             & 41.41 & 15.22 & 13.59 \\
        2 & \checkmark  & \             & \             & 46.64 & 5.63 & 17.33 \\
        3 & \checkmark  & \checkmark    & \             & 46.92 & 4.83 & 17.43 \\
        4 & \           & \             & \checkmark    & 46.29 & 7.68 & 16.82 \\
        \midrule[0.5pt]
        5 & \checkmark  & \checkmark    & \checkmark    & \f{47.23} & \f{3.28} & \f{17.99} \\
        \bottomrule[1.2pt]
    \end{tabular}
    \endgroup
    \caption{Results of different variants of the proposed framework. 
             We use CSP as the baseline model (first row). $\mathit{VS}$ denotes the visual synthesizer module.}
    \label{tab:ablation}
\end{table}


\begin{figure}[t]
    \centering
    \includegraphics[width=1.0\linewidth]{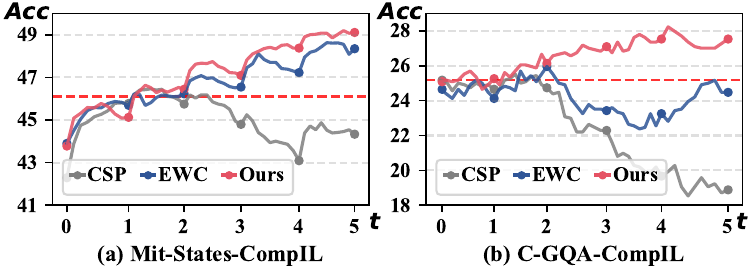}
    \caption{Compositional generalization capability of CSP, equipped with state-of-the-art methods, after training on different tasks in the compositional incremental learning process.The red line indicates CLIP's zero-shot performance.}
    \label{fig:visualize}
\end{figure}

\subsection{Ablation Studies}

To validate the effectiveness of each component, we conduct ablation studies on the MIT-States-CompIL benchmark using CSP as the baseline, with results shown in Table~\ref{tab:ablation}.
Adding the visual synthesizer ($\mathit{VS}$) (row 2) yields significant gains over the baseline (row 1).
Introducing semantic loss ($\mathcal{L}_{sem}$) in row 3 further enhances semantic consistency, improving performance across all metrics.
Linguistic primitive distillation ($\mathcal{L}_{kd}$) also brings improvements, though not as much as the full model.
Combining all components achieves the best overall results, confirming that each module contributes effectively and complementarily.


\subsection{Quantitative Studies}

The accuracy of different continual learning methods on the final task of CompIL throughout the compositional incremental learning process is illustrated in Figure~\ref{fig:visualize}, which reflects the variation in the model's compositional generalization capability as learning new compositions.
We observe that, compared to other methods, our framework significantly enhances the model's compositional generalization capability as it continually learns new compositions.

\section{Conclusion}

In this paper, we have presented a practical and challenging setting for compositional generalization, termed CompIL.
The setting challenges models to continually learn new compositions, aiming to improve their compositional generalization capability progressively.
We have developed a pipeline to construct CompIL benchmarks, resulting in MIT-States-CompIL and C-GQA-CompIL for quantitative evaluation.
Moreover, we have proposed a pseudo-replay framework that can mitigate composition knowledge forgetting and primitive representation drift by leveraging a visual synthesizer and a linguistic primitive distillation mechanism.
Extensive experiments on two CZSL models across the proposed benchmarks demonstrate its effectiveness.

\section{Acknowledgments}

This work was supported by the Shenzhen Science and Technology Program under Grant No. JCYJ20241202130548062, Natural Science Foundation of China (NSFC) under Grants No. 62172041 and No. 62176021, and Natural Science Foundation of Shenzhen under Grant No. JCYJ20230807142703006.

\bibliography{ref}

\newpage
\section{Appendix}

\begin{table*}[h]
    \small
    \centering
    \begingroup
    \setlength{\tabcolsep}{3.5pt}
    \renewcommand{\arraystretch}{0.85}
    \begin{tabular}{l|cccccc|cccccc}
        \toprule[1.2pt]
        \multirow{2.5}{*}{\diagbox[width=0.27\textwidth]{Method}{Benchmark}} & \multicolumn{6}{c|}{MIT-States-CompIL} & \multicolumn{6}{c}{C-GQA-CompIL} \\
        \cmidrule(lr){2-7}\cmidrule(lr){8-13}
        & $\textit{U}$ & $\textit{S}$ & $\textit{AUC}$ & $\textit{mU}$ & $\textit{fS}(\downarrow)$ & $\textit{mAUC}$ & $\textit{U}$ & $\textit{S}$ & $\textit{AUC}$ & $\textit{mU}$ & $\textit{fS}(\downarrow)$ & $\textit{mAUC}$ \\
        \midrule[0.5pt]
        Zero-Shot                           & 41.02 & 28.15 & 8.95 & -    & -     & -     & 23.95 & 6.75 & 1.16 & -    & -     & -     \\
        Joint                               & 45.80 & 42.90 & 15.98 & -    & -     & -     & 25.70 & 27.59 & 5.82 & -    & -     & -     \\
        \midrule[0.5pt]
        Vanilla                             & 34.79 & 25.71 & 7.01 & 36.77 & 14.75 & 10.35 & 14.60 & 15.20 & 1.56 & 17.59 & 11.92 & 3.38 \\
        SI~\cite{zenke2017continual}        & 41.35 & 32.61 & 10.76 & 41.33 & 9.77 & 12.77 & 16.00 & 17.35 & 2.10 & 18.60 & 8.90 & 3.96 \\
        EWC~\cite{chaudhry2019continual}    & \s{44.24} & \s{34.92} & \s{12.35} & \s{42.55} & 7.18 & \s{13.64} & \s{22.90} & 21.00 & \s{3.97} & \s{22.66} & \f{5.15} & \s{5.08} \\
        \midrule[0.5pt]
        A-GEM~\cite{chaudhry2018efficient}  & 41.43 & 32.14 & 10.41 & 40.81 & 11.11 & 12.41 & 13.55 & 15.63 & 1.63 & 17.97 & 10.84 & 3.52 \\
        DER++~\cite{buzzega2020dark}        & 41.34 & 33.61 & 10.97 & 39.20 & 6.19 & 11.96 & 18.41 & 20.16 & 2.82 & 20.05 & 6.12 & 4.23 \\
        \midrule[0.5pt]
        L2P~\cite{wang2022learning}         & 34.09 & 25.92 & 6.79 & 35.37 & 13.96 & 9.46 & 13.90 & 14.54 & 1.53 & 16.73 & 13.22 & 3.11 \\
        AttriCLIP~\cite{wang2023attriclip}  & 33.89 & 28.40 & 7.60 & 36.71 & 13.96 & 10.59 & 15.73 & \s{21.14} & 2.61 & 18.72 & 11.87 & 4.77 \\
        \midrule[0.5pt]
        Ours                                & \f{44.66} & \f{36.98} & \f{13.21} & \f{42.92} & \f{3.91} & \f{14.41} & \f{25.35} & \f{21.52} & \f{4.29} & \f{25.32} & 7.31 & \f{5.32} \\
        \bottomrule[1.2pt]
    \end{tabular}
    \endgroup
    \caption{Comparison with state-of-the-art continual learning methods on the CSP with CLIP-B/16. The best scores are bold.}
    \label{tab:csp-clip-b/16}
\end{table*}

\begin{table*}[h]
    \small
    \centering
    \begingroup
    \setlength{\tabcolsep}{3.5pt}
    \renewcommand{\arraystretch}{0.85}
    \begin{tabular}{l|c|cccccc|cccccc}
        \toprule[1.2pt]
        \multirow{2.5}{*}{\diagbox[width=0.23\textwidth]{Method}{Baseline}} & Buffer & \multicolumn{6}{c|}{CSP~\cite{nayak2023learning} (CLIP ViT-L/14)} & \multicolumn{6}{c}{Troika~\cite{huang2024troika} (CLIP ViT-B/16)} \\
        \cmidrule(lr){3-8}\cmidrule(lr){9-14}
        & Size & $\textit{U}$ & $\textit{S}$ & $\textit{AUC}$ & $\textit{mU}$ & $\textit{fS}(\downarrow)$ & $\textit{mAUC}$ & $\textit{U}$ & $\textit{S}$ & $\textit{AUC}$ & $\textit{mU}$ & $\textit{fS}(\downarrow)$ & $\textit{mAUC}$ \\
        \midrule[0.5pt]
        A-GEM~\cite{chaudhry2018efficient}  & 5\% & 17.66 & 18.69 & 2.53 & 20.45 & 7.06 & 4.36 & 23.51 & 26.83 & 5.01 & 23.90 & 23.52 & 9.11 \\
                                            & 10\% & 18.53 & 17.40 & 2.47 & 20.77 & 7.95 & 4.36 & 24.30 & 26.68 & 4.92 & 24.44 & 24.48 & 9.89 \\
        \midrule[0.5pt]
        DER++~\cite{buzzega2020dark}        & 5\% & 22.64 & 21.24 & 3.59 & 22.40 & 4.25 & 4.64 & 21.33 & 32.85 & 5.84 & 20.00 & 8.32 & 8.45 \\
                                            & 10\% & 22.12 & 19.68 & 3.35 & 21.99 & 4.82 & 4.60 & 23.00 & 32.93 & 6.91 & 22.70 & 12.23 & 9.43 \\
        \bottomrule[1.2pt]
    \end{tabular}
    \endgroup
    \caption{Results of replay-based continual learning methods with different replay buffer sizes on the C-GQA-CompIL benchmark.}
    \label{tab:replay-buffer-c-gqa}
\end{table*}

\subsection{Implementation Details}

We conduct all experiments on a single NVIDIA RTX A40 GPU using Pytorch~\cite{paszke2019pytorch} library.

\subsubsection{Details of Main Paper Figure One}

For each data point in the figure, we sampled three different training sets from MIT-States~\cite{isola2015discovering} and ran under three random seeds for each, resulting in nine experiments per data point. The mean and the 95\% confidence interval (shaded areas) are plotted.

\subsubsection{Details of The Compared Methods}

Since CompIL is a newly proposed setting, there does not exist any prior works that can be used for comparison directly.
To ensure a fair comparison, we reimplement and adapt three types of continual learning methods to comply with CZSL baselines.
Specifically: We built upon the widely adopted Mammoth\footnote{https://github.com/aimagelab/mammoth} codebase, which provides implementations of various continual learning methods, as the foundation of all re-implemented methods.
We also conduct hyperparameter searching based on the best configurations on TinyImagenet~\cite{tiny-imagenet} from Mammoth, to find the optimal configuration for each method.

\subsection{Additional results}

\subsubsection{Results of CSP with The CLIP ViT-B/16 Backbone}

We conduct experiments on CSP with CLIP-B/16 to to examine the effect of the backbone scale and the performance of different CZSL methods with the same backbone.
The experimental results on MIT-States-CompIL and C-GQA-CompIL are listed in Tables~\ref{tab:csp-clip-b/16}, which reveal that:
(1) The main conclusions in the paper still hold: all compared methods underperform the zero-shot baseline on unseen compositions, underscoring the challenges addressed by CompIL.
(2) With a smaller backbone (ViT-B/16 vs. ViT-L/14) reduces the overall performance ceiling, all methods exhibit similar proportional drops. Notably, our method continues to achieve the best results, particularly on unseen compositions. This confirms the robustness and scalability of our framework across backbone sizes, enabling gradual and consistent improvements in compositional generalization capability of models.
(3) Compared to Troika (with the same backbone), more advanced CZSL architectures raise the performance upper bound—especially on the complex dataset, \emph{i.e.}, C-GQA-CompIL. However, these gains come with increased risk of forgetting learned compositions. Our framework effectively mitigates such composition knowledge forgetting and primitive representation drift, and can be integrated with various CZSL methods.

\subsubsection{Replay-based Methods with Different Buffer Sizes}

For the replay-based continual learning methods included in our comparison, we also evaluated the impact of different replay buffer sizes on performance.
Specifically, the buffer size ranged from 5\% to 10\% of the number of samples in the training set.
In contrast, our framework does not require an external replay buffer to store samples.
It leverages a trainable visual synthesizer to continuously synthesize visual representations of different seen compositions.

The experimental results of two CZSL baseline model on C-GQA-ComIL benchmark are shown in the table~\ref{tab:replay-buffer-c-gqa}. 
We observe that increasing the replay buffer size does not lead to consistent performance improvements across metrics for replay-based methods. Specifically, A-GEM and DER++ exhibit decreased compositional generalization performance ($\textit{U}$ and $\textit{mU}$) on the CSP backbone. On the Troika backbone, they show some improvement, but at the cost of a significant increase in forgetting. On the CSP model, both methods fail to improve in terms of $\textit{AUC}$ and $\textit{mAUC}$. These results suggest that simply enlarging the replay buffer is not sufficient to address the challenges in CompIL. It is worth noting that even a 5\% buffer is already relatively large, as the buffer size in class-incremental learning is typically less than 2\%.

\subsubsection{Parameter Analysis of Hyperparameters $\alpha$ and $\beta$}

\begin{figure}[h]
    \centering
    \vspace{-0.35 cm}
    \includegraphics[width=1.0\linewidth]{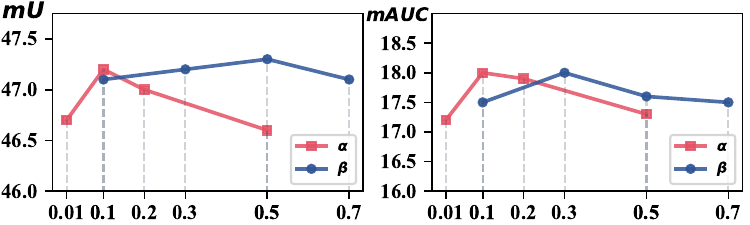}
    \caption{Results on MIT-States-CompIL benchmark under varying values of hyperparameters $\alpha$ and $\beta$.}
    \label{fig:ablation}
\end{figure}

We further analyze the influence of hyperparameters $\alpha$ and $\beta$ with CSP baseline model on MIT-States-CompIL benchmark.
The results in Figure~\ref{fig:ablation} show that $\alpha$ requires careful tuning: low values reduce the semantic consistency of synthesized samples, while too high values hinder VAE training.
In contrast, the framework is relatively robust to variations in $\beta$, with 0.3 yielding the best performance.

\end{document}